\documentclass[sigconf]{acmart}
\renewcommand\footnotetextcopyrightpermission[1]{} 
\usepackage{pgfplots}
\AtBeginDocument{%
  }
\usepackage{subcaption}
\begin{document}

\title{Generative Engine Optimization: A VLM and Agent Framework for Pinterest Acquisition Growth 
}

\author{Faye Zhang}  
\email{fzhang@pinterest.com}  
\affiliation{%
  \institution{Pinterest}
  \country{}
}
\affiliation{%
  \institution{Stanford University}  
  \country{}
} 
\orcid{0009-0005-5801-0206}

\author{Qianyu Cheng}
\email{qcheng@pinterest.com}
\affiliation{%
  \institution{Pinterest}  
  \country{}
} 

\author{Jasmine Wan}  
\email{qwan@pinterest.com}  
\affiliation{%
  \institution{Pinterest}  
  \country{}
} 

\author{Vishwakarma Singh}
\email{vishwakarmasingh@pinterest.com}
\affiliation{%
  \institution{Pinterest}  
  \country{}
}

\author{Jinfeng Rao}
\email{marquisrao@pinterest.com}
\affiliation{%
  \institution{Pinterest} 
  \country{}
} 

\author{Kofi Boakye}
\email{kboakye@pinterest.com}
\affiliation{%
  \institution{Pinterest}  
  \country{}
}




\begin{abstract}
Large Language Models are fundamentally reshaping content discovery through AI-native search systems such as ChatGPT, Gemini, and Claude. Unlike traditional search engines that match keywords to documents, these systems infer user intent, synthesize multimodal evidence, and generate contextual answers directly on the search page, introducing a paradigm shift from Search Engine Optimization (SEO) to Generative Engine Optimization (GEO). For visual content platforms hosting billions of assets, this poses an acute challenge: individual images lack the semantic depth and authority signals that generative search prioritizes, risking disintermediation as user needs are satisfied in-place without site visits.

We present Pinterest GEO, a production-scale framework that pioneers \textit{reverse search design}: rather than generating generic image captions describing what content \textit{is}, we fine-tune Vision-Language Models (VLMs) to predict what users would actually \textit{search for}, augmented this with AI agents that mine real-time internet trends to capture emerging search demand. These VLM-generated queries then drive construction of semantically coherent Collection Pages via multimodal embeddings, creating indexable aggregations optimized for generative retrieval. Finally, we employ hybrid VLM and two-tower ANN architectures to build authority-aware interlinking structures that propagate signals across billions of visual assets. Deployed at scale across billions of images and tens of millions of collections, GEO delivers 20\% organic traffic growth contributing to multi-million monthly active user (MAU) growth, demonstrating a principled pathway for visual platforms to thrive in the generative search era.
\end{abstract}



\begin{teaserfigure}
  \includegraphics[width=\textwidth]{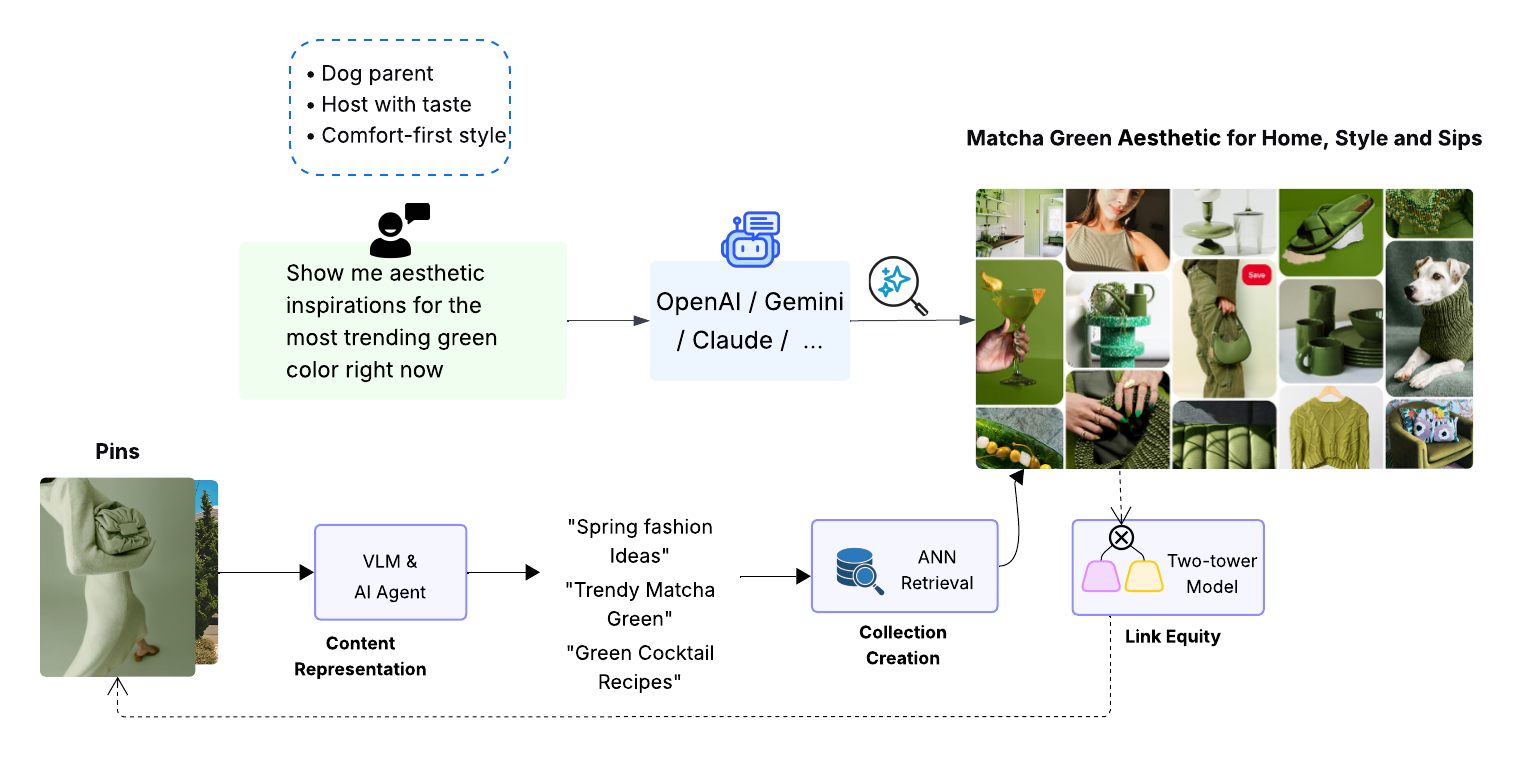}
  \caption{Pinterest GEO Architecture for Content Creation and Publishing, 2025.}
  \label{fig:teaser}
\end{teaserfigure}


\maketitle
\pagestyle{empty}
\section{Introduction}
\label{sec:introduction}

User behavior in web search is undergoing a measurable structural shift. ChatGPT now processes over 1.1 billion queries daily with 800 million weekly active users; Google's AI Overviews reach 2 billion monthly users across 200 countries~\cite{firstpagesage2025,semrush2025}. Referral traffic from AI platforms grew 357\% year-over-year~\cite{seranking2025}. More consequentially, the nature of queries is changing: 57.9\% of searches triggering AI Overviews are phrased as questions, and queries of eight words or longer have a 57\% probability of generating AI-synthesized responses rather than traditional link results~\cite{ahrefs2025}. These systems do not return ranked document lists; they synthesize answers, reason over retrieved evidence, and selectively cite sources judged authoritative~\cite{nakano2021webgpt,menick2022teaching}.

For content platforms dependent on search-driven acquisition, this represents an existential shift. The rules governing visibility have fundamentally changed~\cite{brin1998anatomy,aggrawal2023generative}, yet billions of images, articles, and products remain indexed under assumptions designed for a retrieval paradigm that is rapidly eroding. This transition is particularly acute for visual content~\cite{thomee2016yfcc100m}: unlike text, images lack the lexical surface forms that both traditional and generative search systems require for indexing and authority assessment. 

We formalize this as the visual GEO problem: given a large corpus of images, generate textual representations that (1) align with latent user intent rather than literal visual description, (2) aggregate into coherent topical surfaces that establish citation-worthy authority, and (3) adapt to emerging search demand before it saturates in behavioral logs. This poses three technical challenges: vision-language models~\cite{radford2021learning,liu2023visual} can generate captions (``a woman in a pink dress''), but search optimization requires intent-aligned queries (``garden party outfit ideas''); generative systems preferentially cite consolidated sources over isolated assets~\cite{mao2024search}, yet constructing topically coherent collections at scale has required manual curation~\cite{mcauley2015image}; and search demand is non-stationary~\cite{choi2012predicting}, requiring proactive trend detection before signals appear in behavioral logs.

Existing work addresses these challenges in isolation. VLMs have advanced in caption quality~\cite{li2023blip}, but their deployment for search-oriented generation at billion-scale remains underexplored. Dense retrieval enables semantic matching~\cite{johnson2019billion,malkov2018efficient}, but presupposes query representations exist. Recent GEO theory establishes optimization principles~\cite{aggrawal2023generative}, but focuses on text and lacks production validation on visual corpora.

This paper presents an end-to-end system for visual GEO. We fine-tune VLMs to generate intent-aligned topics using search performance signals, deploy AI agents~\cite{yao2023react} that mine external trend sources~\cite{choi2012predicting} for proactive content creation, construct semantically coherent landing pages via multimodal embeddings~\cite{beal2026pinclip}, and build authority-signaling link structures using hybrid VLM and two-tower ANN architectures~\cite{yi2019sampling}. Deployed across Pinterest's billion-image corpus, the system powers tens of millions of landing pages, delivering 20\% organic traffic growth at 94$\times$ lower inference cost than commercial VLM APIs.

Our contributions are:
\begin{itemize}
    \item Formalization of visual GEO as a distinct problem with analysis of technical barriers differentiating it from text-centric approaches
    \item VLM fine-tuning methodology using search performance signals, achieving 19\% improvement in topic-query alignment over production baselines
    \item AI agent architecture for real-time trend acquisition, enabling content optimization days to weeks ahead of behavioral signals
    \item Production deployment at billion-image scale with ablations quantifying contributions of representation, aggregation, and interlinking
\end{itemize}

\section{Related Work}

\subsection{From SEO to GEO}

Traditional search engines retrieve and rank web pages using indexing and ranking signals~\cite{manning2008introduction}, where SEO optimizes visibility through keywords, metadata, and page structure. GEO represents a fundamental shift: rather than ranking documents, generative engines synthesize answers and selectively cite sources~\cite{aggarwal2024geo}.

Aggarwal et al.~\cite{aggarwal2024geo} formalize this distinction, demonstrating that structured data, citations, and evidence-based annotations significantly improve content selection by generative search engines. We operationalize these principles for visual content through VLM-based annotation and collection construction. Chen et al.~\cite{chen2025role} extend this by emphasizing intent alignment: GEO success requires anticipating diverse user intents rather than optimizing for static queries. We implement this through fine-tuning on search performance signals and AI agent-based trend mining. Mahe Chen et al.~\cite{chen2025dominate} show that generative engines favor "earned media" and domain expertise over traditional on-page SEO factors. We address this by engineering automated interlinking and authority-signaling structures across curated landing surfaces.

Our work extends these GEO principles to visual content at production scale, combining search-optimized representation, semantic aggregation, and authority construction in an end-to-end system.

\subsection{Vision-Language Models for Task-Oriented Image Understanding}

Traditional image captioning optimizes for descriptive accuracy, generating natural language that faithfully describes visible elements~\cite{vinyals2015show}. However, descriptive fidelity does not guarantee task utility to capture search intent or functional context.

Recent work shifts toward task-oriented generation. Instruction-tuned VLMs~\cite{liu2023visual,dai2023instructblip} enable controllable generation through prompts that specify desired reasoning patterns. Visual grounding~\cite{yu2016modeling} and dense captioning~\cite{johnson2016densecap} generate language aligned with spatial attention and user needs rather than exhaustive descriptions.

For search applications, image-to-text retrieval~\cite{chen2020uniter,li2022blip} matches images to relevant textual descriptions. However, these approaches optimize for generic query generation or retrieval similarity rather than search engine performance metrics like click-through rate or ranking position.

Fine-tuning strategies adapt VLMs to specialized objectives. LoRA~\cite{hu2022lora} enables parameter-efficient adaptation, while preference-based learning~\cite{lee2023aligning} incorporates explicit quality signals beyond similarity scores. This work applies intent-driven fine-tuning using search performance signals to train models that generate queries users would actually issue, bridging the gap between visual description and search intent.

\subsection{Representation Learning and Semantic Retrieval for Content Aggregation}

Semantic retrieval systems embed high-dimensional content into metric spaces where similarity reflects task-relevant relationships~\cite{bengio2013representation}. Vision-language models like CLIP~\cite{radford2021learning} enable zero-shot cross-modal retrieval by aligning image and text encoders through contrastive learning on web-scale data. However, foundation model embeddings optimize for broad semantic coverage rather than domain-specific objectives like user engagement or purchase intent~\cite{muennighoff2023mteb}.

Task-specific fine-tuning substantially improves retrieval quality for specialized domains~\cite{gao2021simcse}. Two-tower architectures~\cite{yi2019sampling} enable efficient production deployment by encoding queries and items independently, allowing offline indexing while incorporating behavioral signals (clicks, saves, conversions) during training. However, embedding objectives create fundamental tradeoffs: visual similarity embeddings maintain aesthetic coherence but miss functional relationships~\cite{bell2015learning}; engagement-optimized embeddings improve conversion but may sacrifice semantic relevance~\cite{covington2016deep}.

Production systems deploy approximate nearest neighbor (ANN) algorithms for sub-linear retrieval over billion-scale corpora. HNSW graphs~\cite{malkov2018efficient} achieve logarithmic query complexity through multi-layer proximity structures, dominating industry deployments~\cite{johnson2019billion}.

\subsection{Link Equity for Search Optimization}

Link equity, formalized through PageRank~\cite{page1999pagerank}, quantifies how hyperlink structures distribute authority across web content. Internal linking architectures leverage this mechanism to signal semantic relationships and topical coherence to search engines: well-linked pages receive higher crawl priority, stronger ranking signals, and improved indexation~\cite{linkvector2024,serpforge2024}. Traditional SEO exploits this through hub-and-spoke structures where authoritative landing pages link to related content, distributing ranking potential while creating navigable topic hierarchies~\cite{brin1998anatomy}.

In the generative search era, link equity maintains relevance but shifts function. Rather than merely improving ranking position, structured link topologies help LLMs interpret entity relationships and content coherence~\cite{aggarwal2024geo}. Generative systems preferentially cite well-connected, contextually rich content surfaces over isolated pages: consolidated sources with clear semantic structure provide interpretable evidence chains that support answer generation~\cite{nakano2021webgpt,menick2022teaching}. Recent work demonstrates that content with explicit internal linking to related entities achieves higher citation rates in AI-generated responses~\cite{aggarwal2024geo}.

For visual content, systematic link construction faces unique challenges: images lack textual anchors and native hyperlink structures. Prior work addresses visual retrieval through ANN systems~\cite{johnson2019billion,malkov2018efficient} and multimodal embeddings~\cite{radford2021learning}, but focuses on query-time retrieval rather than authority-building link structures. This work bridges the gap by engineering systematic internal linking between visual assets and topically coherent landing pages, operationalizing link equity principles for image-centric platforms in both traditional and generative search contexts.

\section{Methodology}

\subsection{Step 1: Content Representation via VLM and AI Agents}
\label{sec:vlm}

\begin{figure}[h]
    \centering
    \includegraphics[width=1\linewidth]{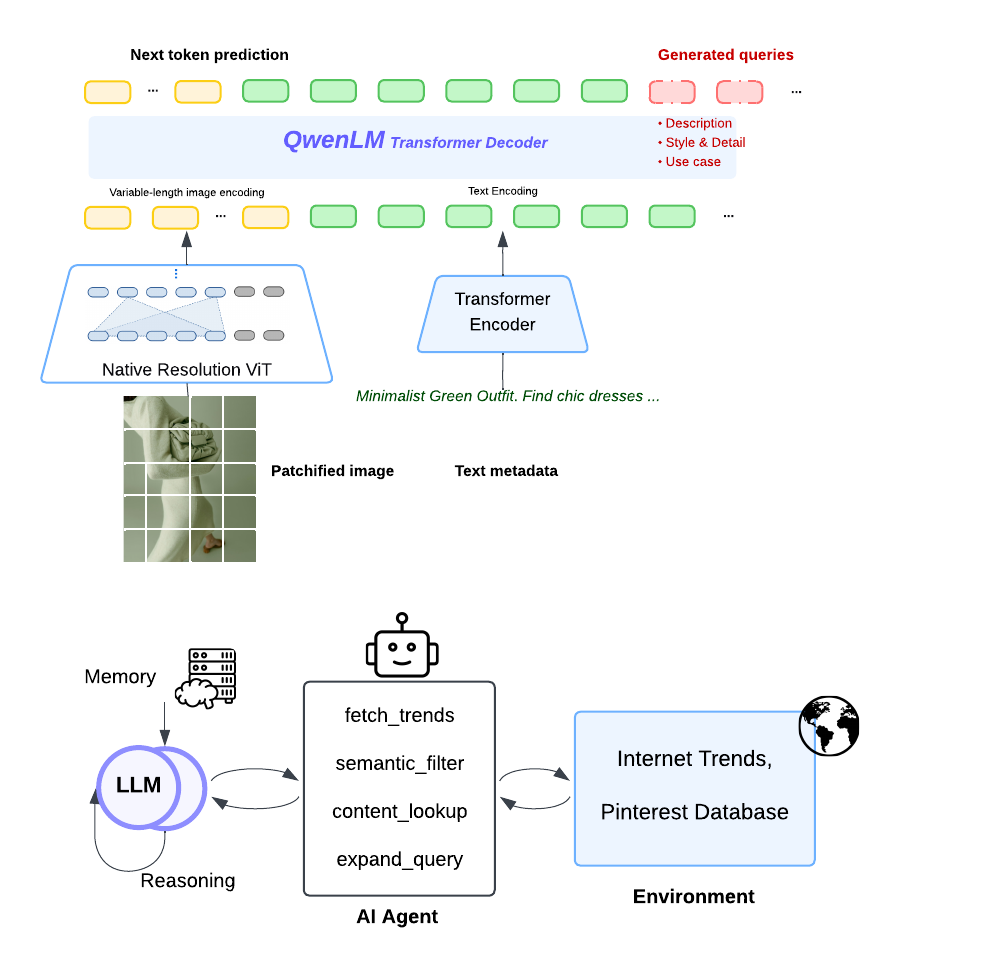}
    \caption{VLM and AI Agent Framework to Generate Content Topics}
    \label{fig:vlm}
\end{figure}

The content representation stage transforms visual assets into search-optimized textual annotations. We develop two complementary systems: (1) a supervised fine-tuned VLM that generates intent-aligned queries from images, and (2) an AI agent framework that mines real-time internet trends to proactively create content for emerging search demand. (Figure \ref{fig:vlm})

\subsubsection{VLM Supervised Fine-Tuning}
\label{sec:vlm_training}

\paragraph{Task Formulation}

Given an image $x$ and optional contextual metadata $c$ (e.g., board title, Pin description, creator category), our objective is to generate a set of textual queries $\mathcal{Q} = \{q_1, q_2, \ldots, q_k\}$ that maximize the probability of the image being retrieved and cited by search systems. This differs fundamentally from image captioning~\cite{vinyals2015show,chen2015microsoft}: rather than describing visual content, we aim to predict the queries users would issue when seeking this content.

We decompose the output space into three query categories based on analysis of high-performing search terms:

\begin{itemize}
    \item \textbf{Description queries (30\%)}: Entity-level identifications (``Soft Green Knit Dress'') targeting users searching for specific subjects.
    
    \item \textbf{Style and detail queries (30\%)}: Attribute-augmented variations capturing visual aesthetics, materials, colors, and design specifics (``Sage Green Monochrome Look'') to capture refinement intent.
    
    \item \textbf{Use case queries (40\%)}: Intent-oriented phrases describing applications, occasions, or contexts (``Capsule Wardrobe Must-Haves 2026,'' ``Modern Office Outfits for Women'). These queries capture latent user goals that purely descriptive annotations miss.
\end{itemize}

The 30/30/40 distribution reflects empirical analysis showing use-case queries drive disproportionate incremental traffic, as they address intent not captured by existing description-based annotation systems. 

\paragraph{Model Architecture}

We build upon Qwen2-VL-7B-Instruct~\cite{qwen2024}, a pretrained vision-language model~\cite{radford2021learning} supporting dynamic resolution image inputs and strong multilingual capabilities across our target markets. The model architecture comprises:

\begin{itemize}
    \item \textbf{Vision encoder}: Leveraging the native dynamic resolution support \cite{native_dynamic}, a Vision Transformer (ViT) \cite{vit} is modified to process images of any resolution. This allows the model to dynamically generate a variable number of visual tokens for each image, e.g., a 224 × 224 image is compressed into 66 tokens before being passed to the language decoder; 
    \item \textbf{Language decoder}: 7B parameter transformer~\cite{vaswani2017attention} generating text autoregressively, conditioned on fused visual-textual representations.
\end{itemize}

We apply parameter-efficient fine-tuning via Low-Rank Adaptation (LoRA)~\cite{hu2022lora}, updating only low-rank decomposition matrices while freezing pretrained weights. This reduces trainable parameters by $>$99\% while preserving the model's general visual understanding capabilities.

\paragraph{Training Data Construction}

Training data quality is critical given the divergence from standard captioning objectives. We construct approximately 100K training examples (5--10K held out for evaluation) through a multi-stage pipeline combining behavioral signals with synthetic augmentation.

\textbf{Stage 1: Mining search performance signals.} We extract query-image associations from external search engine console, which provides external search engine metrics for Pinterest URLs. For each image signature, we retrieve associated queries and filter for demonstrated search performance:
\begin{equation}
\text{retain}(q, x) = \begin{cases}
\text{True} & \text{if } \texttt{impressions}(q, x) > 1000 \\
\text{True} & \text{if } \texttt{impressions}(q, x) > 10 \,\land \\
& \qquad \texttt{CTR}(q, x) \geq 0.8 \\
\text{True} & \text{if } \texttt{impressions}(q, x) > 10 \,\land \\
& \qquad \texttt{position}(q, x) \leq 10 \\
\text{False} & \text{otherwise}
\end{cases}
\end{equation}

where $\texttt{position}$ is the average ranking in search results (1-indexed, lower is better). We select the top 30 queries per image signature ranked by impressions and position, yielding query-image pairs grounded in demonstrated user demand.

\textbf{Stage 2: Synthetic augmentation} Search console data skews toward description and style queries that already perform well; use-case queries are underrepresented because they often lack historical coverage. To address this cold-start problem, we generate 200K synthetic examples using GPT-4V~\cite{openai2023gpt4} as an oracle labeler.

Given an image, we prompt GPT-4V to generate candidate queries across all three categories, conditioned on:
\begin{itemize}
    \item Visual content analysis (objects, scenes, styles, colors)
    \item Inferred user intent (what would someone searching for this want to accomplish?)
    \item Pinterest-specific context (DIY projects, inspiration boards, shopping intent)
\end{itemize}

We also apply classification to label each query by category: description, style/detail, or use case. This enables stratified sampling during training to achieve the target 30/30/40 distribution, ensuring the model learns to generate all query types.

\paragraph{Training Configuration}
We trained Qwen2-VL-7B-Instruct using supervised fine-tuning (SFT)~\cite{wei2022finetuned} formulated as
conditional language modeling. Each example consists of conditioning inputs—
image(s) $x$, optional context metadata $c$, and an optional task identifier
$t$—together with a prompt $p$ and a ground-truth target response sequence
$y=(y_1,\ldots,y_{|y|})$ (e.g., tags, queries, JSON, or free-form text). The
training objective minimizes the negative log-likelihood of the target response
tokens:
\begin{equation}
\mathcal{L}(\theta)
= - \sum_{(x,c,t,p,y)\in \mathcal{D}}
\sum_{i=1}^{|y|}
\log p_{\theta}\!\left(y_i \mid y_{<i}, x, c, t, p\right).
\label{eq:sft_clm}
\end{equation}
In instruction/message tuning, we compute loss only on the assistant/response
tokens $y$ by masking system/user/prompt tokens in the labels. 

Training executes on p4d.24xlarge instances (8$\times$A100 80GB GPUs) via TCP. Key hyperparameters:

\begin{table}[h]
\centering
\small
\begin{tabular}{ll}
\toprule
\textbf{Parameter} & \textbf{Value} \\
\midrule
Base model & Qwen2-VL-7B-Instruct \\
Fine-tuning method & LoRA \\
Batch size (per device) & 2 \\
Gradient accumulation & 8 steps \\
Learning rate & 2e-5 (cosine decay) \\
Training epochs & 1--3 \\
Max image pixels & 602,112 \\
Max sequence length & 1,024 tokens \\
Evaluation strategy & Per epoch \\
\bottomrule
\end{tabular}
\caption{VLM training hyperparameters.}
\label{tab:hyperparams}
\end{table}

\paragraph{Inference Pipeline}

At inference time, we generate 5--7 queries per image across the three categories. The pipeline processes images through several stages:

\textbf{Input preparation.} Images are fetched from Pinterest's CDN at 736x resolution (the most reliable variant) and preprocessed to the model's expected format. We cap resolution at 602,112 pixels (min 175,616) to balance detail preservation with inference efficiency.

\textbf{Generation.} We use vLLM~\cite{kwon2023efficient}-based batch inference with the following decoding parameters:
\begin{itemize}
    \item Temperature: 0.1 (low for high precision)
    \item Top-p: 0.001 (near-greedy sampling)
    \item Top-k: 1
    \item Repetition penalty: 1.05
    \item Max new tokens: 256
\end{itemize}

These conservative parameters prioritize precision over diversity, as downstream deduplication and fusion with ANN-based candidates provides sufficient coverage.

\textbf{Post-processing.} Raw VLM outputs undergo multi-stage validation:

\begin{enumerate}
    \item \textbf{Parsing}: Extract structured query outputs from model responses, handling JSON formatting and category tags.
    
    \item \textbf{Safety filtering}: Standard Pinterest safety text layer removes explicit content~\cite{wulczyn2017ex}.
    
    \item \textbf{LLM Validation}: A fine-tuned LLaMA-7B~\cite{touvron2023llama} classifier evaluates each query for topic eligibility, checking:
    \begin{itemize}
        \item Brand safety and trademark compliance
        \item Pinterest voice alignment (inspirational, positive framing)
        \item Search intent strength 
        \item Grey-zone filtering (content that passes safety filter but exhibits negative sentiment, implicit bias, or borderline appropriateness)
    \end{itemize}
    
    \item \textbf{EAI framework}: Pinterest's Empathetic AI evaluation ensures outputs meet fairness and inclusivity guidelines.
    
    \item \textbf{Language classification}: Queries are tagged by language and routed to locale-appropriate downstream pipelines.
    
    \item \textbf{Deduplication}: Embedding-based semantic similarity~\cite{reimers2019sentence} identifies and merges near-duplicate queries.
\end{enumerate}

\subsubsection{Agentic Trend Mining}
\label{sec:agents}

VLM-generated annotations operate reactively on existing content, unable to anticipate emerging search demand that has yet to materialize in platform behavioral logs. We address this cold-start problem through an autonomous agent system that proactively discovers and exploits nascent search trends.

\paragraph{Problem Formulation}

Given external trend signal streams $\mathcal{S}_t = \{s_1, s_2, \ldots, s_n\}$ at time $t$ (e.g., Google Trends queries~\cite{choi2012predicting}), we seek to generate a set of Pinterest-relevant query expansions $\mathcal{Q}_{\text{trend}} = \{q_1, q_2, \ldots, q_m\}$ where each $q_i$ satisfies: (1) semantic alignment with platform taxonomy, (2) content sufficiency (retrievable Pins $>$ threshold $\tau$), and (3) temporal relevance (trend lifecycle stage conducive to content creation latency).

\paragraph{Agent Architecture}

We implement a ReAct-style~\cite{yao2023react} agent using LangGraph~\cite{langgraph} for orchestration, enabling iterative reasoning and tool use through cyclic state transitions.

\textbf{State Representation:} The agent maintains structured state $\mathcal{M} = (\mathcal{M}_{\text{short}}, \mathcal{M}_{\text{long}})$ comprising:
\begin{itemize}
    \item $\mathcal{M}_{\text{short}}$: Episode memory storing current session context (active trends, intermediate reasoning, tool outputs)
    \item $\mathcal{M}_{\text{long}}$: Persistent memory indexing historical trend performance, seasonal patterns, and category-specific conversion signals
\end{itemize}

\textbf{Tool Interface:} The agent executes through function calling~\cite{schick2024toolformer} to external systems:
\begin{itemize}
    \item \texttt{fetch\_trends(region, timespan)}: Retrieves trending queries via external data sources with temporal and geographic filtering
    \item \texttt{semantic\_filter(trend, threshold)}: LLM-based classifier predicting Pinterest relevance $p(r|t) \in [0,1]$
    \item \texttt{content\_lookup(query)}: Queries ANN index~\cite{malkov2018efficient} to verify Pin availability, returns count and quality scores
    \item \texttt{expand\_query(trend, taxonomy)}: Generates category-conditioned expansions using few-shot prompting~\cite{brown2020language} with exemplars from $\mathcal{M}_{\text{long}}$
\end{itemize}

\textbf{Orchestration Flow:} The agent executes via a directed acyclic graph (DAG) of nodes representing reasoning steps:
\begin{enumerate}
    \item \textbf{Planning node}: LLM generates execution strategy conditioned on weekly schedule and market priorities
    \item \textbf{Retrieval node}: Parallel tool calls fetch trends across 
    \item \textbf{Filtering node}: Batch classification filters trends using learned relevance function, rejecting low-fit categories (news, sports, politics)
    \item \textbf{Expansion node}: Conditional generation produces taxonomy-aligned query variants for retained trends
    \item \textbf{Validation node}: Routes generated queries to shared post-processing pipeline (Section~\ref{sec:vlm_training})
\end{enumerate}

State transitions follow: $s_{t+1} = f(s_t, a_t, o_t)$ where $a_t$ is the tool action, $o_t$ is the observed output, and $f$ updates both short-term and long-term memory.

\subsection{Step 2: Content Collection Generation}
VLM-generated queries define search intent; the next challenge is constructing collections that aggregate semantically relevant Pins. At billion-image scale, this requires ANN retrieval for sub-linear query time. Collection quality depends critically on embedding architecture: different training objectives produce distinct retrieval behaviors and engagement outcomes.

We leverage Manas~\cite{pinterest2024manas}, Pinterest's HNSW-based ANN system, with two embedding architectures: PinCLIP~\cite{beal2026pinclip}, optimizing for visual-semantic coherence via co-save signals, and SearchSAGE~\cite{pinterest2021searchsage}, optimizing for engagement via query-click data and entity-graph context. We evaluate both to determine optimal collection construction strategy.

\subsubsection{PinCLIP: Multimodal Pin Representation}

PinCLIP\cite{pinclip} encodes each Pin using both its cover image and descriptive text. The image $I_j$ and text $T_j$ of Pin $j$ are encoded via separate transformer encoders, $E_\mathrm{img}$ and $E_\mathrm{txt}$, then merged by a transformer aggregator $E_\mathrm{agg}$ to yield a Pin embedding $\mathbf{m}_j = E_\mathrm{agg}(E_\mathrm{img}(I_j), E_\mathrm{txt}(T_j)) \in \mathbb{R}^d$.

\begin{figure}[h]
    \centering
    \includegraphics[width=1\linewidth]{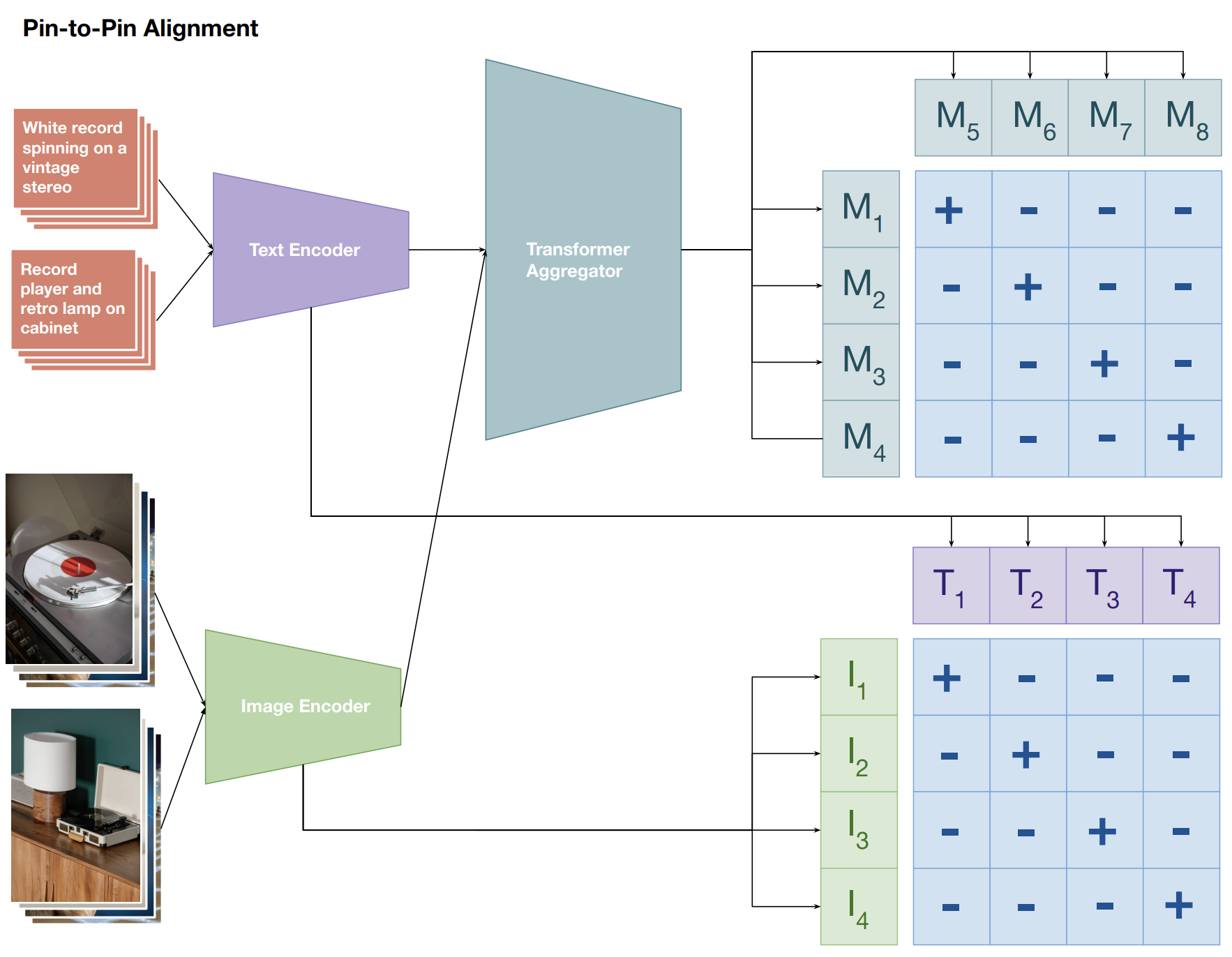}
    \caption{PinCLIP: Pin-to-Pin Loss}
    \label{fig:pinclip}
\end{figure}

This multimodal Pin embedding is trained via the sum of two softmax losses:
\begin{itemize}
    \item \textbf{Image-Text Loss}: aligns each Pin's image embedding $\mathbf{x}_i$ with its text embedding $\mathbf{y}_j$;
    \item \textbf{Pin-Pin Loss}: aligns Pin embeddings $\mathbf{x}_i, \mathbf{y}_j$ for Pin pairs saved on the same Pinterest board, encouraging higher similarity among contextually related Pins. (Figure \ref{fig:pinclip})
\end{itemize}
For a positive pair $(i, j)$ and negative pool of size $|\mathcal{C}|$, the loss has the general softmax form
$$
\mathcal{L}_{ij} = -\log \frac{\exp(\mathbf{x}_i^\top \mathbf{y}_j / \tau)}{\sum_{k \in \mathcal{C}} \exp(\mathbf{x}_i^\top \mathbf{y}_k / \tau)}
$$
where $\tau$ is a temperature hyperparameter. The total objective sums the two losses:
$$
\mathcal{L}_{\mathrm{PinCLIP}} = \mathbb{E}_{(i, j)\,\in\,\mathcal{P}_\mathrm{img-txt}}[\mathcal{L}_{ij}]
+ \mathbb{E}_{(i,j)\,\in\,\mathcal{P}_\mathrm{Pin-Pin}}[\mathcal{L}_{ij}]
$$
where $\mathcal{P}_\mathrm{img-txt}$ and $\mathcal{P}_\mathrm{Pin-Pin}$ are batches of positive image-text and Pin-Pin pairs, respectively.

\subsubsection{SearchSAGE: Multi-Entity, Graph-Aware Representation}

SearchSAGE represents queries and Pinterest catalog entities (Pins, products, boards, etc.) using text transformer encoders. Each entity $E_j$ and each query $Q_i$ are mapped to vectors $\mathbf{e}_j = E_\mathrm{ent}(E_j)$ and $\mathbf{q}_i = E_\mathrm{qry}(Q_i)$ in $\mathbb{R}^d$. These encoders are augmented with graph-based context using Pinterest’s internal entity graph, which captures relations among users, Pins, products, and boards. (Figure \ref{fig:searchsage})

Training uses positive pairs $(Q_i, E_j)$ based on user engagement (e.g., from saves or clicks), with a separate softmax loss for each type of relation (such as \emph{query-Pin}, \emph{query-product}):
$$
\mathcal{L}_{ij}^T = -\log \frac{\exp(\mathbf{q}_i^\top \mathbf{e}_j / \tau)}{\sum_{k \in \mathcal{C}} \exp(\mathbf{q}_i^\top \mathbf{e}_k / \tau)}
$$
where $T$ is the task type, and $\mathcal{C}$ is a set of negative samples for $E_j$. The total SearchSAGE loss sums over all task types:
$$
\mathcal{L}_\mathrm{SearchSAGE} = \sum_{T \in \mathcal{T}}\,\mathbb{E}_{(i, j) \in\,T}[\mathcal{L}_{ij}^T]
$$

\begin{figure}
    \centering
    \includegraphics[width=1\linewidth]{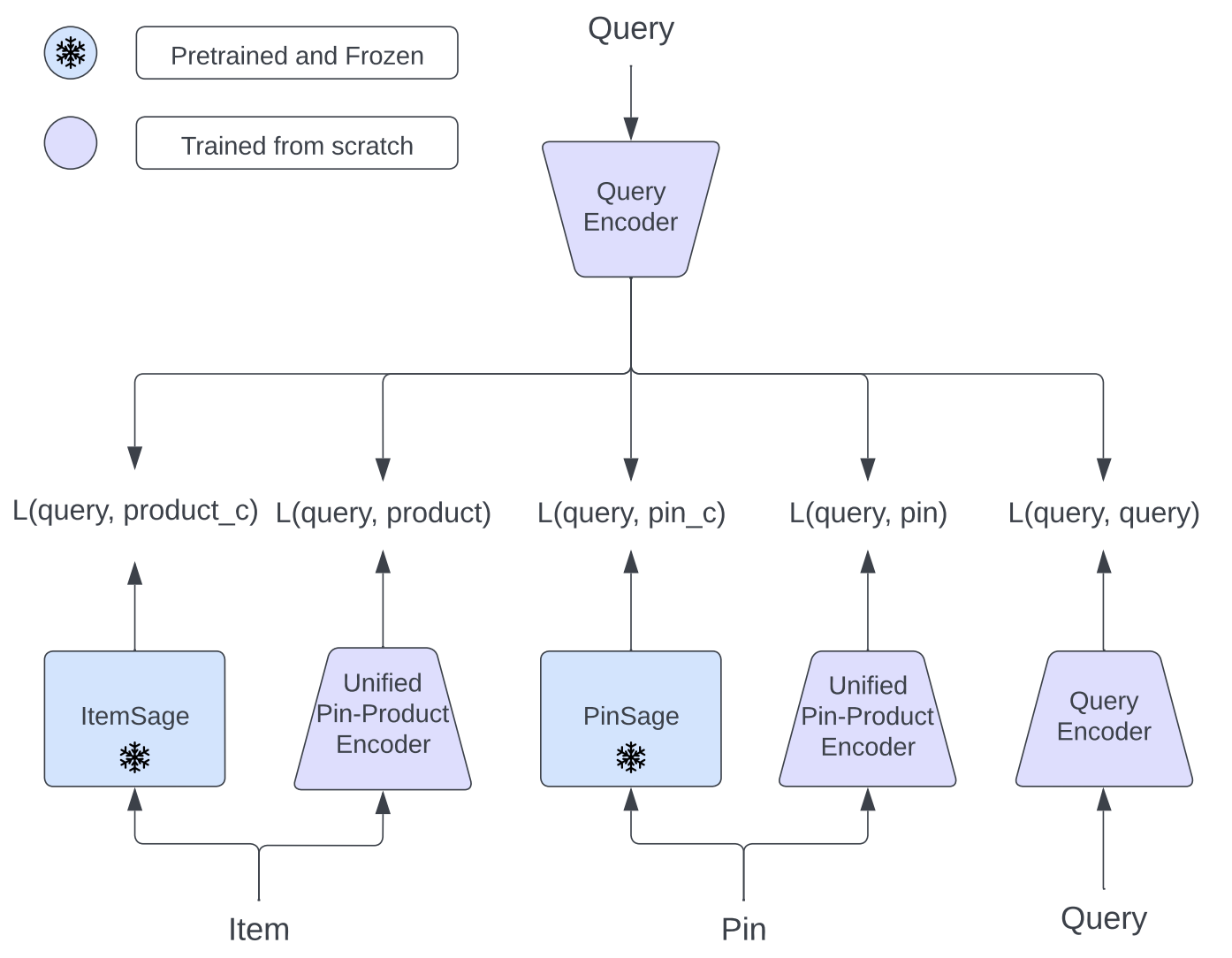}
    \caption{SearchSAGE: Multi-Entity, Graph-Aware Architecture}
    \label{fig:searchsage}
\end{figure}

\subsubsection{Manas: HNSW-based ANN Framework}

To perform large-scale content retrieval using the embeddings from PinCLIP and SearchSAGE, we utilize Manas~\cite{pinterest2024manas}, Pinterest's internal ANN infrastructure. Manas is based on the HNSW algorithm, an efficient approach for identifying approximate nearest neighbors in large, high-dimensional vector spaces.

In practice, we index all Pin embeddings within Manas, building an ANN index that supports real-time, scalable search. At retrieval time, each topic or user query is encoded, then used to probe the Manas index, which efficiently returns the most relevant Pins according to vector similarity. This framework enables low-latency collection generation across Pinterest's massive and ever-evolving content catalog.

\subsection{Step 3: Content Distribution}

To maximize content discoverability, we systematically annotate Pins with VLM-generated queries and construct internal link structures that propagate authority signals. The Visual Annotation for Search Engine (VASE) system ranks and selects the most contextually relevant queries for each Pin. Annotations serve dual purposes: they provide textual metadata enabling search engine indexation of visual content, and they create hyperlinks to Content Collection \cite{pinlanding2025}, constructing the link equity topology.

\textbf{Model Architecture}
Each tower consists of a deep multi-layer perceptron (MLP) with three fully connected layers (layer sizes,[512, 384, 256]), interleaved with ReLU activations, Layer Normalization, and Dropout for robust generalization. 

The Pin Tower processes the Pin’s visual, textual, and metadata features, including a 1028-dim Unifying Visual Embeddings \cite{pinterest2021uve}, a 768-dim GPT-generated text embedding, and a 1-dim image Perception Score. 

The Query Tower takes the candidate query’s features, consisting of a 768-dim SEO VASE GPT embedding and a 1-dim query length normalization score. 

Both towers project to a 128-dim embedding and are L2 normalized before the final similarity calculation.
\begin{figure}
    \centering
    \includegraphics[width=1\linewidth]{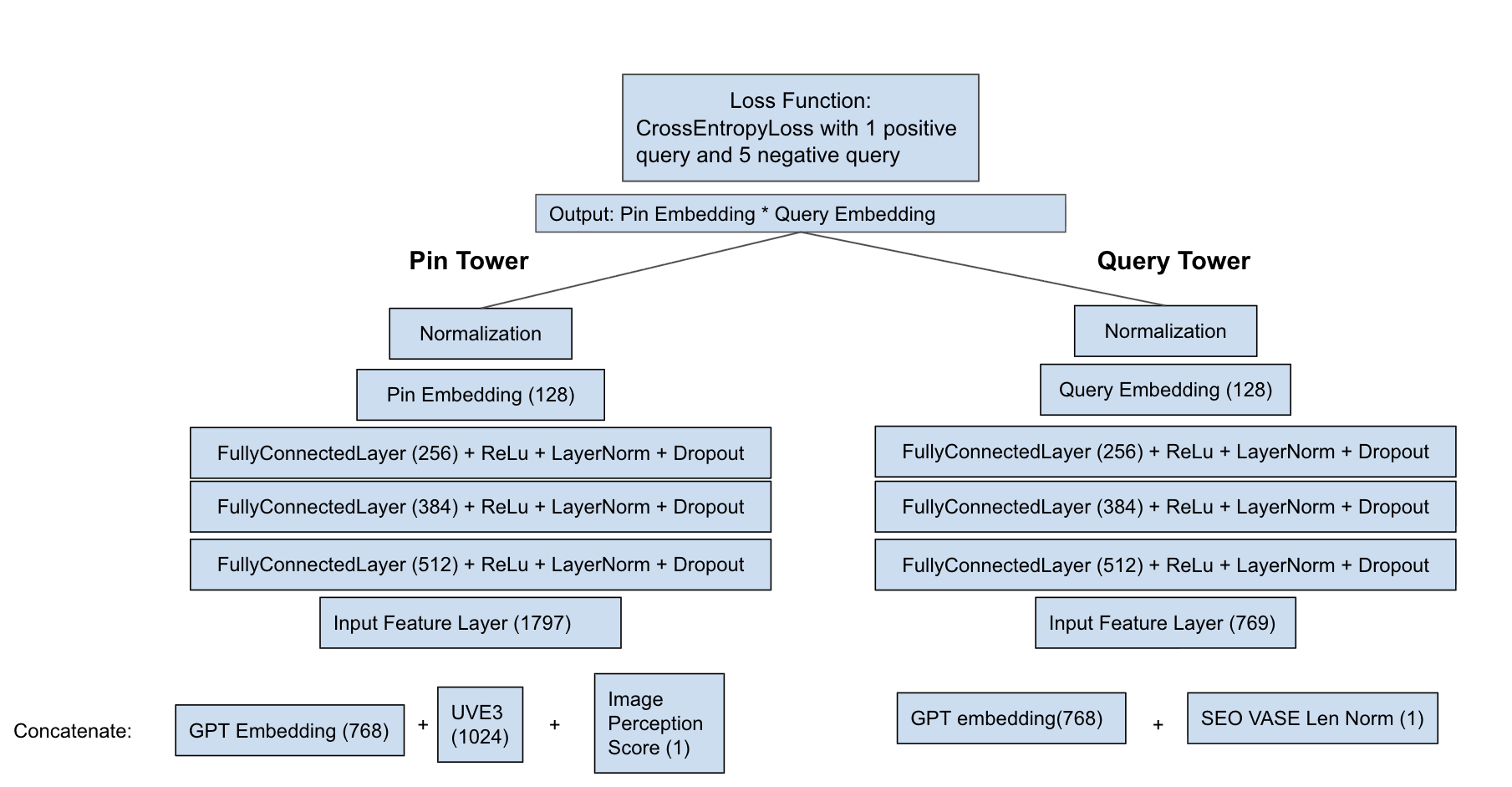}
    \caption{VASE: Two Tower Ranking Model}
    \label{fig:searchsage}
\end{figure}

\textbf{Model Input Features}
The input features to the Pin tower and query tower are as follows: 
\begin{itemize} 
\item \textbf{Unified Visual Embeddings (UEv3)}: High-dimensional vectors representing each image, generated by deep neural networks trained to capture visual features, style, and content semantics.
\item \textbf{Image Perception Score}: A regression-based score derived from a dedicated MLP model trained on large-scale human-rated image quality datasets, predicting perceived visual quality from the UEv3 image embeddings. 
\item \textbf{GPT Text Embeddings}: Dense text representations produced by large language models (e.g., GPT-2), pre-trained and fine-tuned on rich Pinterest data, including Pin titles, descriptions, queries, comments, ads, and board text, to capture nuanced textual and semantic signals. 
\end{itemize} By concatenating these diverse feature sets from across the Pinterest ecosystem, the model attains a holistic understanding of each Pin—encompassing visual, textual, and metadata attributes.

\textbf{Loss Function} \ The model is trained using a Margin Ranking Loss (also known as triplet margin or contrastive margin loss), which encourages the Pin embedding to be closer to positive query embeddings and farther from negative ones by a specified margin. Formally, given a Pin embedding $E_{Pin}$, a positive query embedding $E_{pos}$, and a negative query embedding $E_{neg}$, the loss for each sample is computed as:
\begin{equation}
\mathcal{L}_{\text{margin}} = \max\left(0, \ E_{\mathrm{Pin}} \cdot E_{\mathrm{neg}} - E_{\mathrm{Pin}} \cdot E_{\mathrm{pos}} + m\right)
\end{equation}
where $m$ is the margin hyperparameter (set to 0.95 in our implementation), and $E_{Pin} \cdot E_{query}$ denotes the dot product (or cosine similarity if normalized) between the Pin and query embeddings.

This loss structure encourages the model to assign a higher similarity score to true Pin-query pairs (positive) than to randomly sampled negative pairs, enforcing discriminative separation by at least the margin.(e.g., margin=0.95 yields optimal separation: eval correct rank 0.981 and eval loss 0.045). 

During training, the model learns to maximize the similarity between Pins and semantically correct queries, while minimizing similarity with negatives, selected either from hard competitors or random candidates. Post-training, the final scoring is done via a dot product/cosine similarity of normalized embeddings.

\textbf{Labeling Strategy}
Our training targets are derived from an extensive synthetic labeling pipeline that combines performance data from Google and internal Pinterest search with semantic overlap and relevance heuristics. For each \textit{signature-annotation} pair, labels are assigned as follows:

\begin{itemize} \item \textbf{Positive (label = 1):} Assigned if the Pin has demonstrated good search performance (from Google or internal logs) and the associated query shows strong semantic match

\item \textbf{Negative, hard (label = -1):} These negatives are randomly drawn from other pairs of signature-annotation pair that are not semantically related.  \end{itemize}

To refine label confidence, any signature-annotation pair with high navboost coverage ($>$0.54) is promoted based on manual review thresholds. Navboost provides keyword/phrase candidates based on actual navigation and engagement signals (i.e., terms that users have demonstrated interest in by interacting with search results, feeds, or content)

The synthetic labeling pipeline draws data from one year of Google performance logs, daily-flattened internal search, sampling from both high-traffic and SEO-referred signatures. This ensures robust model generalization across head, torso, and tail segments and across major supported languages.

\section{Experiments and Results}

\subsection{VLM Training Evaluation}
\label{sec:vlm_eval}

We evaluate the supervised fine-tuned VLM through three complementary methodologies: automated n-gram overlap metrics, LLM-based semantic assessment, and human expert annotation. The evaluation is conducted on a held-out test set of 1,800 image signatures across diverse visual categories and query types.

\subsubsection{Automated Metric Evaluation}

We compute ROUGE-1 F1 scores~\cite{lin2004rouge} to measure token-level overlap between generated queries and ground-truth annotations. Table~\ref{tab:rouge_scores} reports performance stratified by query category.

\begin{table}[h]
\centering
\begin{tabular}{lc}
\toprule
\textbf{Query Category} & \textbf{ROUGE-1 F1} \\
\midrule
Image Descriptions & 0.63 \\
Search Modifiers (Style/Detail) & 0.40 \\
Use Case Queries & 0.34 \\
\midrule
Overall & 0.46 \\
\bottomrule
\end{tabular}
\caption{ROUGE-1 F1 scores by query category on held-out test set (N=1,800). Lower scores for search modifiers and use cases reflect higher semantic variance in these categories compared to factual descriptions.}
\label{tab:rouge_scores}
\end{table}

The descending performance from descriptions (0.63) to use cases (0.34) aligns with increasing semantic diversity and abstraction levels. Image descriptions converge toward canonical entity names (e.g., "mid-century modern living room"), yielding higher lexical overlap. Use case queries exhibit greater paraphrastic variation (e.g., "college dorm bedroom ideas" vs. "how to decorate a student room"), reducing n-gram precision while maintaining semantic equivalence.

\subsubsection{LLM-Based Semantic Evaluation}

To capture semantic correctness beyond surface-form matching, we employ GPT-4o as an automated evaluator. For each test example, we provide the image, concatenated metadata (board title, Pin description, creator category), model-generated queries, and ground-truth labels. The evaluator assesses five dimensions on binary scales:

\begin{itemize}
    \item \textbf{Relevance}: Query semantically matches image content
    \item \textbf{Specificity}: Query contains sufficient detail for targeted retrieval
    \item \textbf{Category Fit}: Query aligns with Pinterest taxonomy (fashion, home, food, etc.)
    \item \textbf{Diversity}: Query set covers multiple user intents
    \item \textbf{Coverage}: Generated queries span description, style, and use case categories
\end{itemize}

Table~\ref{tab:gpt_eval} compares model outputs against ground-truth labels. The VLM achieves $>$93\% across all metrics, with minimal degradation relative to human-annotated targets. The 1--2\% performance gap is attributable to GPT-4o's bias toward its own synthetic training data, which constitutes a portion of the ground-truth set.

\begin{table}[h]
\centering
\begin{tabular}{lccc}
\toprule
\textbf{Metric} & \textbf{Model Output} & \textbf{Ground Truth} & \textbf{$\Delta$} \\
\midrule
Relevance & 96.15\% & 97.70\% & -1.55\% \\
Specificity & 93.41\% & 94.59\% & -1.18\% \\
Category Fit & 96.41\% & 97.70\% & -1.29\% \\
Diversity & 97.47\% & 99.18\% & -1.71\% \\
Coverage & 96.84\% & 99.18\% & -2.34\% \\
\bottomrule
\end{tabular}
\caption{GPT-4o semantic evaluation on held-out test set (N=1,800). Ground truth includes GPT-4o synthetic annotations, introducing favorable bias toward labels.}
\label{tab:gpt_eval}
\end{table}

\subsubsection{Human Expert Evaluation}

We conduct A/B comparative evaluation between VLM-generated annotations and production baseline (VASE ANN retrieval system) using 140 randomly sampled image signatures. Three expert annotators rate outputs on 5-point Likert scales for relevance and diversity.

\paragraph{Evaluation Protocol}

Annotators are shown: (1) the source image, (2) VLM-generated queries (5 queries), and (3) production baseline queries (up to 9 queries from ANN retrieval). They independently rate each system on:
\begin{itemize}
    \item \textbf{Relevance} (1--5): Semantic alignment between queries and image content
    \item \textbf{Diversity} (1--5): Coverage of distinct user intents and search contexts
\end{itemize}

For VLM outputs specifically, annotators additionally assess use case accuracy: whether generated use case queries plausibly match user search intent.

\paragraph{Results}

Table~\ref{tab:human_eval} summarizes comparative ratings. VLM outputs achieve 19\% higher relevance (4.47 vs. 3.28, $p < 0.01$, paired t-test) while maintaining competitive diversity despite generating fewer queries (5 vs. 9). The production baseline's diversity advantage stems from ANN retrieval covering broader semantic neighborhoods, though at the cost of reduced precision.

The use case query, the primary innovation differentiating our approach from standard captioning, achieves 4.15 accuracy in human evaluation compared to 2.21 baseline, demonstrating that the model successfully learns to generate actionable search intents beyond visual descriptions. 

\begin{table}[h]
\centering
\begin{tabular}{lcc}
\toprule
\textbf{Metric (1--5)} & \textbf{VLM} & \textbf{Production Baseline} \\
\midrule
Relevance & \textbf{4.47} & 3.28 \\
Diversity & 3.43 & \textbf{3.54} \\
Use Case Accuracy & \textbf{4.15} & 2.21 \\
\midrule
Query Count & 5 & 9 \\
\bottomrule
\end{tabular}
\caption{Human expert evaluation comparing VLM-generated queries against production ANN baseline (N=140 images, 3 annotators).}
\label{tab:human_eval}
\end{table}

\subsection{Embedding Model Comparison: SearchSAGE vs. PinCLIP}

To evaluate and compare the content quality of collections retrieved by PinCLIP and SearchSAGE, we conducted both offline and online experiments:

\subsubsection{Offline Evaluation}
For each topic query, we retrieved the top 10 Pins using each kind of embeddings. We then designed a GPT-based evaluator, which analyzes the cover image and descriptive text for each Pin in the returned collection, and judge whether each Pin can satisfied the intent of the topic query. The \emph{query intent satisfying rate} is computed as the percentage of retrieved Pins deemed relevant.

\subsubsection{Online A/B Testing}
We deployed collections generated by PinCLIP and SearchSAGE in an A/B test over one month. For each variant, we measured:
\begin{itemize}
    \item \textbf{Signup rate:} The percentage of sessions where a user signed up after clicking into a collection page.
    \item \textbf{Login rate:} The percentage of sessions where a user logged in after clicking into a collection page.
    \item \textbf{Clickthrough rate (CTR):} The percentage of sessions in which a user clicked through at least one Pin in the collection.
\end{itemize}

\begin{table}[h]
  \centering
  \caption{Comparison of quality metrics for collections retrieved by PinCLIP and SearchSAGE.}
  \label{tab:quality_comparison}
  \begin{tabular}{lcc}
    \toprule
    \textbf{Metric} & \textbf{PinCLIP} & \textbf{SearchSAGE} \\
    \midrule
    Intent satisfying rate (top 10, offline) & \textbf{0.881} & 0.848 \\
    Signup rate (\%)                              & 0.121   & \textbf{0.127} \\
    Login rate (\%)                               & \textbf{0.365}   & 0.361 \\
    Clickthrough rate (\%)                        & 1.61    & \textbf{1.66} \\
    \bottomrule
  \end{tabular}
\end{table}

\subsubsection{Result}
As shown in Table~\ref{tab:quality_comparison}, PinCLIP achieves a higher offline query intent satisfying rate, indicating better semantic alignment with the topic query as assessed by large language model judgment. However, SearchSAGE outperforms PinCLIP in online signup rate and clickthrough rate. This advantage is likely attributable to SearchSAGE's training method, which optimizes for user engagement by leveraging engagement-driven and graph-based signals. In summary, while PinCLIP excels at intent matching, SearchSAGE demonstrates superior effectiveness at driving real user interactions.

\subsection{Link Equity Ablation and Production Study}

We deployed VLM-generated annotations in large-scale A/B
testing to validate their impact on traffic and discoverability. We evaluate the contribution of systematic annotation and interlinking through three conditions: (1) \textbf{Ablation}: no VASE system, Pins lack annotations and interlinks; (2) \textbf{Control}: ANN-retrieved annotations with metadata-based linking; (3) \textbf{VLM}: VLM-generated annotations with intent-aligned linking.

\begin{table}[h]
  \centering
  \caption{Traffic impact across annotation strategies over 4-week A/B test.}
  \label{tab:vlm_traffic_lift}
  \begin{tabular}{lccc}
    \toprule
    \textbf{Metric} & \textbf{Ablation } & \textbf{Control} & \textbf{Enabled} \\
    \midrule
     User Sessions         & 0.82$\times$ & 1.0$\times$ & 1.18$\times$ \\
    \bottomrule
  \end{tabular}
\end{table}

\begin{table}[h]
  \centering
  \caption{Generative search traffic overrepresentation by annotation strategy.}
  \label{tab:geo_overrep}
  \begin{tabular}{lc}
    \toprule
    \textbf{Experiment Groups} & \textbf{GEO Traffic Multiplier} \\
    \midrule
    Ablation  & 0.04 $\times$  \\
    Control       & 1.0$\times$  \\
    Enabled       & 9.2$\times$ \\
    \bottomrule
  \end{tabular}
\end{table}

\paragraph{Traffic Results}
Over four weeks, VLM-enabled Pins demonstrated substantial traffic gains (Table~\ref{tab:vlm_traffic_lift}). The ablation demonstrates that systematic annotation provides 18\% traffic gain over unlinked Pins. VLM annotations deliver an additional 18\% improvement over ANN baselines, attributable to superior intent alignment and semantic quality.

For generative engine specially, VLM-annotated content received 9.2$\times$ more user traffic compared to 1$\times$ for ANN-matched content (Table~\ref{tab:geo_overrep}). This confirms that intent-aligned VLM annotations improve visibility in both traditional and generative search environments.

\paragraph{Scalability via ANN + Two-Tower Ranking}
While VLM inference achieves superior quality, billion-scale deployment requires efficient candidate generation. We leverage HNSW-based ANN retrieval with PinCLIP/SearchSAGE embeddings to match images to candidate queries, followed by lightweight two-tower MLP ranking for final selection. 
This hybrid approach drove a 20\% production traffic lift vs. control, with 94$\times$ lower inference cost than commercial VLM APIs and no degradation in engagement.(Table~\ref{tab:perc_diff_two_tower}).

\begin{table}[h]
  \centering
  \caption{User engagement lift from ANN + two-tower annotation system.}
  \label{tab:perc_diff_two_tower}
  \begin{tabular}{lc}
    \toprule
    \textbf{Metric} & \textbf{Lift (\%)} \\
    \midrule
    Closeups           & +1.24 \\
    Clickthrough       & +1.20 \\
    Search             & +0.94 \\
    RePin              & +1.10 \\
    Total Sessions     & +1.20 \\
    Signup Success     & +0.83 \\
    Login Success      & +1.83 \\
    \bottomrule
  \end{tabular}
\end{table}

\section{Future Directions}
\label{sec:future}

While our current system demonstrates substantial production impact, several promising research directions remain unexplored.

\paragraph{Reinforcement Learning from Search Engine Feedback}

Our supervised fine-tuning approach relies on historical search console data and human annotations, both of which lag emerging search engine preferences and algorithmic updates. A natural extension is reinforcement learning~\cite{ouyang2022training,rafailov2024direct} to directly optimize for search engine outcomes: indexation, bot-crawl, impression, click-through rate, ranking position, and AI-citation. The key challenge lies in the opacity and delay of reward signals: content crawl, indexation, and ranking occur over days to weeks, and the causal mechanisms remain largely black-box. Techniques such as offline RL~\cite{levine2020offline}, policy gradient methods~\cite{schulman2017proximal}, or preference-based learning~\cite{rafailov2024direct} could enable continuous adaptation to evolving search algorithms without requiring manual data curation.

\paragraph{Causal Discovery for Search Ranking}

Search engine ranking remains largely a black box, especially across different AI search engines. Causal inference methods~\cite{pearl2009causality,scholkopf2021toward} could help disentangle which content features (query semantics, link structure, visual quality) causally drive indexation and ranking versus mere correlations, enabling more principled optimization strategies.

\section{Conclusion}
\label{sec:conclusion}

The rise of AI-native search systems fundamentally transforms how visual content is discovered and surfaced on the web. We present Pinterest GEO, a production-scale framework that aligns visual content platforms with LLM-driven retrieval through three innovations: reverse search VLM design that generates actionable user intents rather than descriptions, high-precision content aggregation via multimodal embeddings, and hybrid VLM-ANN architectures for billion-scale authority construction.

Deployed across hundreds of millions of images, GEO delivers measurable impact: 20\% increase in organic traffic, improved indexation, and enhanced AI-search visibility. Our work demonstrates that adapting to generative search requires fundamentally rethinking content representation and distribution: moving beyond classical SEO toward AI-first design. The framework provides a generalizable blueprint for visual platforms navigating this transition.


\clearpage 
\bibliographystyle{ACM-Reference-Format}
\bibliography{bibliography}

@article{brin1998anatomy,
  title={The anatomy of a large-scale hypertextual web search engine},
  author={Brin, Sergey and Page, Lawrence},
  journal={Computer networks and ISDN systems},
  volume={30},
  number={1-7},
  pages={107--117},
  year={1998}
}

@inproceedings{thomee2016yfcc100m,
  title={YFCC100M: The new data in multimedia research},
  author={Thomee, Bart and Shamma, David A and Friedland, Gerald and Elizalde, Benjamin and Ni, Karl and Poland, Douglas and Borth, Damian and Li, Li-Jia},
  booktitle={Communications of the ACM},
  volume={59},
  number={2},
  pages={64--73},
  year={2016}
}

@inproceedings{mcauley2015image,
  title={Image-based recommendations on styles and substitutes},
  author={McAuley, Julian and Targett, Christopher and Shi, Qinfeng and Van Den Hengel, Anton},
  booktitle={Proceedings of the 38th international ACM SIGIR conference on research and development in information retrieval},
  pages={43--52},
  year={2015}
}

@inproceedings{li2023blip,
  title={BLIP-2: Bootstrapping language-image pre-training with frozen image encoders and large language models},
  author={Li, Junnan and Li, Dongxu and Savarese, Silvio and Hoi, Steven},
  booktitle={International conference on machine learning},
  pages={19730--19742},
  year={2023},
  organization={PMLR}
}

@article{ouyang2022training,
  title={Training language models to follow instructions with human feedback},
  author={Ouyang, Long and Wu, Jeffrey and Jiang, Xu and Almeida, Diogo and Wainwright, Carroll and Mishkin, Pamela and Zhang, Chong and Agarwal, Sandhini and Slama, Katarina and Ray, Alex and others},
  journal={Advances in Neural Information Processing Systems},
  volume={35},
  pages={27730--27744},
  year={2022}
}

@article{rafailov2024direct,
  title={Direct preference optimization: Your language model is secretly a reward model},
  author={Rafailov, Rafael and Sharma, Archit and Mitchell, Eric and Manning, Christopher D and Ermon, Stefano and Finn, Chelsea},
  journal={Advances in Neural Information Processing Systems},
  volume={36},
  year={2024}
}

@inproceedings{schulman2017proximal,
  title={Proximal policy optimization algorithms},
  author={Schulman, John and Wolski, Filip and Dhariwal, Prafulla and Radford, Alec and Klimov, Oleg},
  journal={arXiv preprint arXiv:1707.06347},
  year={2017}
}

@article{levine2020offline,
  title={Offline reinforcement learning: Tutorial, review, and perspectives on open problems},
  author={Levine, Sergey and Kumar, Aviral and Tucker, George and Fu, Justin},
  journal={arXiv preprint arXiv:2005.01643},
  year={2020}
}

@book{pearl2009causality,
  title={Causality},
  author={Pearl, Judea},
  year={2009},
  publisher={Cambridge university press}
}

@article{scholkopf2021toward,
  title={Toward causal representation learning},
  author={Sch{\"o}lkopf, Bernhard and Locatello, Francesco and Bauer, Stefan and Ke, Nan Rosemary and Kalchbrenner, Nal and Goyal, Anirudh and Bengio, Yoshua},
  journal={Proceedings of the IEEE},
  volume={109},
  number={5},
  pages={612--634},
  year={2021},
  publisher={IEEE}
}

@article{schick2024toolformer,
  title={Toolformer: Language Models Can Teach Themselves to Use Tools},
  author={Schick, Timo and Dwivedi-Yu, Jane and Dess{\`i}, Roberto and Raileanu, Roberta and Lomeli, Maria and Zettlemoyer, Luke and Cancedda, Nicola and Scialom, Thomas},
  journal={Advances in Neural Information Processing Systems},
  volume={36},
  year={2023}
}

@article{yao2023react,
  title={ReAct: Synergizing Reasoning and Acting in Language Models},
  author={Yao, Shunyu and Zhao, Jeffrey and Yu, Dian and Du, Nan and Shafran, Izhak and Narasimhan, Karthik and Cao, Yuan},
  journal={arXiv preprint arXiv:2210.03629},
  year={2023}
}

@misc{langgraph,
  title={LangGraph: Building stateful, multi-actor applications with LLMs},
  author={LangChain},
  year={2024},
  howpublished={\url{https://github.com/langchain-ai/langgraph}}
}

@article{hu2022lora,
  title={LoRA: Low-Rank Adaptation of Large Language Models},
  author={Hu, Edward J and Yelong, Shen and Wallis, Phillip and Allen-Zhu, Zeyuan and Li, Yuanzhi and Wang, Shean and Wang, Lu and Chen, Weizhu},
  journal={arXiv preprint arXiv:2106.09685},
  year={2022}
}

@article{qwen2024,
  title={Qwen2-VL: Enhancing Vision-Language Model's Perception of the World at Any Resolution},
  author={Wang, Peng and Bai, Shuai and Tan, Sinan and Wang, Shijie and Fan, Zhihao and Bai, Jinze and Chen, Keqin and Liu, Xuejing and Wang, Jialin and Ge, Wenbin and Fan, Yang and Dang, Kai and Du, Mengfei and Ren, Xuancheng and Men, Rui and Liu, Dayiheng and Zhou, Chang and Zhou, Jingren and Lin, Junyang},
  journal={arXiv preprint arXiv:2409.12191},
  year={2024}
}

@inproceedings{vinyals2015show,
  title={Show and tell: A neural image caption generator},
  author={Vinyals, Oriol and Toshev, Alexander and Bengio, Samy and Erhan, Dumitru},
  booktitle={Proceedings of the IEEE conference on computer vision and pattern recognition},
  pages={3156--3164},
  year={2015}
}

@inproceedings{chen2015microsoft,
  title={Microsoft coco captions: Data collection and evaluation server},
  author={Chen, Xinlei and Fang, Hao and Lin, Tsung-Yi and Vedantam, Ramakrishna and Gupta, Saurabh and Doll{\'a}r, Piotr and Zitnick, C Lawrence},
  booktitle={arXiv preprint arXiv:1504.00325},
  year={2015}
}

@article{vaswani2017attention,
  title={Attention is all you need},
  author={Vaswani, Ashish and Shazeer, Noam and Parmar, Niki and Uszkoreit, Jakob and Jones, Llion and Gomez, Aidan N and Kaiser, {\L}ukasz and Polosukhin, Illia},
  journal={Advances in neural information processing systems},
  volume={30},
  year={2017}
}

@article{openai2023gpt4,
  title={GPT-4 technical report},
  author={OpenAI},
  journal={arXiv preprint arXiv:2303.08774},
  year={2023}
}

@article{wei2022finetuned,
  title={Finetuned language models are zero-shot learners},
  author={Wei, Jason and Bosma, Maarten and Zhao, Vincent Y and Guu, Kelvin and Yu, Adams Wei and Lester, Brian and Du, Nan and Dai, Andrew M and Le, Quoc V},
  journal={arXiv preprint arXiv:2109.01652},
  year={2022}
}

@article{kwon2023efficient,
  title={Efficient memory management for large language model serving with pagedattention},
  author={Kwon, Woosuk and Li, Zhuohan and Zhuang, Siyuan and Sheng, Ying and Zheng, Lianmin and Yu, Cody Hao and Gonzalez, Joseph E and Zhang, Hao and Stoica, Ion},
  journal={arXiv preprint arXiv:2309.06180},
  year={2023}
}

@inproceedings{wulczyn2017ex,
  title={Ex machina: Personal attacks seen at scale},
  author={Wulczyn, Ellery and Thain, Nithum and Dixon, Lucas},
  booktitle={Proceedings of the 26th international conference on world wide web},
  pages={1391--1399},
  year={2017}
}

@inproceedings{reimers2019sentence,
  title={Sentence-bert: Sentence embeddings using siamese bert-networks},
  author={Reimers, Nils and Gurevych, Iryna},
  booktitle={Proceedings of the 2019 Conference on Empirical Methods in Natural Language Processing and the 9th International Joint Conference on Natural Language Processing (EMNLP-IJCNLP)},
  pages={3982--3992},
  year={2019}
}

@article{choi2012predicting,
  title={Predicting the present with Google Trends},
  author={Choi, Hyunyoung and Varian, Hal},
  journal={Economic record},
  volume={88},
  pages={2--9},
  year={2012}
}

@article{malkov2018efficient,
  title={Efficient and robust approximate nearest neighbor search using hierarchical navigable small world graphs},
  author={Malkov, Yu A and Yashunin, Dmitry A},
  journal={IEEE transactions on pattern analysis and machine intelligence},
  volume={42},
  number={4},
  pages={824--836},
  year={2018},
  publisher={IEEE}
}

@inproceedings{lin2004rouge,
  title={ROUGE: A package for automatic evaluation of summaries},
  author={Lin, Chin-Yew},
  booktitle={Text summarization branches out},
  pages={74--81},
  year={2004}
}

@inproceedings{brown2020language,
  title={Language models are few-shot learners},
  author={Brown, Tom and Mann, Benjamin and Ryder, Nick and Subbiah, Melanie and Kaplan, Jared D and Dhariwal, Prafulla and Neelakantan, Arvind and Shyam, Pranav and Sastry, Girish and Askell, Amanda and others},
  booktitle={Advances in Neural Information Processing Systems},
  volume={33},
  pages={1877--1901},
  year={2020}
}

@article{touvron2023llama,
  title={Llama 2: Open foundation and fine-tuned chat models},
  author={Touvron, Hugo and Martin, Louis and Stone, Kevin and Albert, Peter and Almahairi, Amjad and Babaei, Yasmine and Bashlykov, Nikolay and Batra, Soumya and Bhargava, Prajjwal and Bhosale, Shruti and others},
  journal={arXiv preprint arXiv:2307.09288},
  year={2023}
}

@article{aggrawal2023generative,
  title={Generative engine optimization},
  author={Aggrawal, Pranjal and Verma, Vishwas and Doras, Avanika and Narasimhan, Karthik and Nakov, Preslav},
  journal={arXiv preprint arXiv:2311.09735},
  year={2023}
}

@inproceedings{mao2024search,
  title={From search engines to generative engines: A paradigm shift in information retrieval},
  author={Mao, Yuning and He, Pengcheng and Liu, Xiaodong and Shen, Yelong and Gao, Jianfeng and Han, Jiawei and Chen, Weizhu},
  booktitle={Proceedings of the 47th International ACM SIGIR Conference on Research and Development in Information Retrieval},
  pages={2156--2166},
  year={2024}
}

@article{nakano2021webgpt,
  title={WebGPT: Browser-assisted question-answering with human feedback},
  author={Nakano, Reiichiro and Hilton, Jacob and Balaji, Suchir and Wu, Jeff and Ouyang, Long and Kim, Christina and Hesse, Christopher and Jain, Shantanu and Kosaraju, Vineet and Saunders, William and others},
  journal={arXiv preprint arXiv:2112.09332},
  year={2021}
}

@article{menick2022teaching,
  title={Teaching language models to support answers with verified quotes},
  author={Menick, Jacob and Trebacz, Maja and Mikulik, Vladimir and Aslanides, John and Song, Francis and Chadwick, Martin and Glaese, Mia and Young, Susannah and Campbell-Gillingham, Lucy and Irving, Geoffrey and others},
  journal={arXiv preprint arXiv:2203.11147},
  year={2022}
}

@inproceedings{radford2021learning,
  title={Learning transferable visual models from natural language supervision},
  author={Radford, Alec and Kim, Jong Wook and Hallacy, Chris and Ramesh, Aditya and Goh, Gabriel and Agarwal, Sandhini and Sastry, Girish and Askell, Amanda and Mishkin, Pamela and Clark, Jack and others},
  booktitle={International Conference on Machine Learning},
  pages={8748--8763},
  year={2021},
  organization={PMLR}
}

@inproceedings{liu2023visual,
  title={Visual instruction tuning},
  author={Liu, Haotian and Li, Chunyuan and Wu, Qingyang and Lee, Yong Jae},
  booktitle={Advances in Neural Information Processing Systems},
  volume={36},
  pages={34892--34916},
  year={2023}
}

@inproceedings{beal2026pinclip,
  title={PinCLIP: Large-scale Foundational Multimodal Representation at Pinterest},
  author={Beal, Josh and Kim, Eric and Rao, Jinfeng and Wu, Rex and Kislyuk, Dmitry and Rosenberg, Charles},
  booktitle={Proceedings of The ACM Web Conference 2026},
  series={TheWebConf '26},
  year={2026},
  publisher={ACM},
  address={New York, NY, USA},
  pages={9},
  url={https://www.pinterestcareers.com/media/eoqd5wcs/pinclip.pdf}
}

@inproceedings{yi2019sampling,
  title={Sampling-bias-corrected neural modeling for large corpus item recommendations},
  author={Yi, Xinyang and Yang, Ji and Hong, Lichan and Cheng, Derek Zhiyuan and Heldt, Lukasz and Kumthekar, Aditee and Zhao, Zhe and Wei, Li and Chi, Ed},
  booktitle={Proceedings of the 13th ACM Conference on Recommender Systems},
  pages={269--277},
  year={2019}
}

@article{johnson2019billion,
  title={Billion-scale similarity search with GPUs},
  author={Johnson, Jeff and Douze, Matthijs and J{\'e}gou, Herv{\'e}},
  journal={IEEE Transactions on Big Data},
  volume={7},
  number={3},
  pages={535--547},
  year={2019},
  publisher={IEEE}
}

@inproceedings{aggarwal2024geo,
  author    = {Pranjal Aggarwal and Vishvak Murahari and Tanmay Rajpurohit and Ashwin Kalyan and Karthik Narasimhan and Ameet Deshpande},
  title     = {{Geo: Generative Engine Optimization}},
  booktitle = {Proceedings of the 30th ACM SIGKDD Conference on Knowledge Discovery and Data Mining},
  pages     = {5--16},
  year      = {2024}
}

@book{manning2008introduction,
  author    = {Christopher D. Manning},
  title     = {Introduction to Information Retrieval},
  publisher = {Syngress Publishing},
  year      = {2008}
}

@article{chen2025role,
  author    = {Xiaolu Chen and Haojie Wu and Jie Bao and Zhen Chen and Yong Liao and Hu Huang},
  title     = {Role-augmented intent-driven generative search engine optimization},
  journal   = {arXiv preprint arXiv:2508.11158},
  year      = {2025}
}

@misc{firstpagesage2025,
  author       = {{First Page Sage}},
  title        = {Top Generative {AI} Chatbots by Market Share -- {December} 2025},
  howpublished = {\url{https://firstpagesage.com/reports/top-generative-ai-chatbots/}},
  year         = {2025},
  note         = {Accessed: December 2025}
}

@misc{semrush2025,
  author       = {{Semrush}},
  title        = {{AI} Overviews Study: What 2025 {SEO} Data Tells Us About {Google's} Search Shift},
  howpublished = {\url{https://www.semrush.com/blog/semrush-ai-overviews-study/}},
  year         = {2025},
  note         = {Analysis of 10M+ keywords, January--November 2025}
}

@misc{seranking2025,
  author       = {{SE Ranking}},
  title        = {{AI} Traffic in 2025: Comparing {ChatGPT}, {Perplexity} and Other Top Platforms},
  howpublished = {\url{https://seranking.com/blog/ai-traffic-research-study/}},
  year         = {2025},
  note         = {Accessed: September 2025}
}

@misc{ahrefs2025,
  author       = {{Ahrefs}},
  title        = {{AI} Overviews Study: Query Patterns and Citation Analysis},
  howpublished = {\url{https://ahrefs.com/blog/}},
  year         = {2025},
  note         = {November 2025. Reports 57.9\% question queries, 46\% long-tail queries triggering AI Overviews}
}

@article{gemini,
  title={Gemini: a family of highly capable multimodal models},
  author={Team, Gemini and Anil, Rohan and Borgeaud, Sebastian and Alayrac, Jean-Baptiste and Yu, Jiahui and Soricut, Radu and Schalkwyk, Johan and Dai, Andrew M and Hauth, Anja and Millican, Katie and others},
  journal={arXiv preprint arXiv:2312.11805},
  year={2023}
}

@article{pinclip,
  title={PinCLIP: Large-scale Foundational Multimodal Representation at Pinterest},
  author={Beal, Josh and Kim, Eric and Rao, Jinfeng and Wu, Rex and Kislyuk, Dmitry and Rosenberg, Charles},
  year={2026}
}

@inproceedings{searchsage,
  title={OmniSearchSage: Multi-Task Multi-Entity Embeddings for Pinterest Search},
  author={Agarwal, Prabhat and Sk, Minhazul Islam and Pancha, Nikil and Hazra, Kurchi Subhra and Xu, Jiajing and Rosenberg, Chuck},
  booktitle={Companion Proceedings of the ACM Web Conference 2024},
  pages={121--130},
  year={2024}
}

@article{native_dynamic,
  title={Patch n’pack: Navit, a vision transformer for any aspect ratio and resolution},
  author={Dehghani, Mostafa and Mustafa, Basil and Djolonga, Josip and Heek, Jonathan and Minderer, Matthias and Caron, Mathilde and Steiner, Andreas and Puigcerver, Joan and Geirhos, Robert and Alabdulmohsin, Ibrahim M and others},
  journal={Advances in Neural Information Processing Systems},
  volume={36},
  pages={2252--2274},
  year={2023}
}

@article{vit,
  title={An image is worth 16x16 words: Transformers for image recognition at scale},
  author={Dosovitskiy, Alexey},
  journal={arXiv preprint arXiv:2010.11929},
  year={2020}
}

@article{chen2025dominate,
  author    = {Mahe Chen and Xiaoxuan Wang and Kaiwen Chen and Nick Koudas},
  title     = {Generative Engine Optimization: How to Dominate AI Search},
  journal   = {arXiv preprint arXiv:2509.08919},
  year      = {2025},
  url       = {https://arxiv.org/abs/2509.08919}
}

@misc{linkvector2024,
  author = {LinkVector.io},
  title = {Internal Linking Case Study: How Adding Internal Links Improved Ranking for 83\% of Orphan Pages},
  year = {2024},
  note = {Available at: \url{https://linkvector.io/internal-linking-case-study-increase-in-ranking-319}},
}

@misc{serpforge2024,
  author = {SERPForge.io},
  title = {SEO Study: Internal Link Distribution and Its Impact on Traffic},
  year = {2024},
  note = {Available at: \url{https://serpforge.io/link-buildings/link-equity/}},
}

@misc{pinterest2021uve,
  author = {Junnan Li and Weidi Xie and Igor Milyutin and Yanjun Qi and Ravi Rajagopal and Zhiyuan Shi and Priya Goyal},
  title = {Unifying Visual Embeddings for Visual Search at Pinterest},
  howpublished = {\url{https://medium.com/pinterest-engineering/unifying-visual-embeddings-for-visual-search-at-pinterest-74ea7ea103f0}},
  year = {2021},
  note = {Pinterest Engineering Blog}
}

@article{page1999pagerank,
  title={The PageRank citation ranking: Bringing order to the web},
  author={Page, Lawrence and Brin, Sergey and Motwani, Rajeev and Winograd, Terry},
  journal={Technical Report, Stanford University},
  year={1999}
}

@article{bengio2013representation,
  title={Representation learning: A review and new perspectives},
  author={Bengio, Yoshua and Courville, Aaron and Vincent, Pascal},
  journal={IEEE transactions on pattern analysis and machine intelligence},
  volume={35},
  number={8},
  pages={1798--1828},
  year={2013},
  publisher={IEEE}
}

@article{muennighoff2023mteb,
  title={MTEB: Massive text embedding benchmark},
  author={Muennighoff, Niklas and Tazi, Nouamane and Magne, Lo{\"\i}c and Reimers, Nils},
  journal={arXiv preprint arXiv:2210.07316},
  year={2022}
}

@inproceedings{bell2015learning,
  title={Learning visual similarity for product design with convolutional neural networks},
  author={Bell, Sean and Bala, Kavita},
  booktitle={ACM transactions on graphics (TOG)},
  volume={34},
  number={4},
  pages={1--10},
  year={2015},
  organization={ACM}
}

@inproceedings{covington2016deep,
  title={Deep neural networks for youtube recommendations},
  author={Covington, Paul and Adams, Jay and Sargin, Emre},
  booktitle={Proceedings of the 10th ACM conference on recommender systems},
  pages={191--198},
  year={2016},
  organization={ACM}
}

@inproceedings{gao2021simcse,
  title={SimCSE: Simple contrastive learning of sentence embeddings},
  author={Gao, Tianyu and Yao, Xingcheng and Chen, Danqi},
  booktitle={Proceedings of the 2021 Conference on Empirical Methods in Natural Language Processing},
  pages={6894--6910},
  year={2021}
}

@article{dai2023instructblip,
  title={InstructBLIP: Towards general-purpose vision-language models with instruction tuning},
  author={Dai, Wenliang and Li, Junnan and Li, Dongxu and Tiong, Anthony Meng Huat and Zhao, Junqi and Wang, Weisheng and Li, Boyang and Fung, Pascale and Hoi, Steven},
  journal={arXiv preprint arXiv:2305.06500},
  year={2023}
}

@inproceedings{yu2016modeling,
  title={Modeling context in referring expressions},
  author={Yu, Licheng and Poirson, Patrick and Yang, Shan and Berg, Alexander C and Berg, Tamara L},
  booktitle={European Conference on Computer Vision},
  pages={69--85},
  year={2016},
  organization={Springer}
}

@inproceedings{johnson2016densecap,
  title={Densecap: Fully convolutional localization networks for dense captioning},
  author={Johnson, Justin and Karpathy, Andrej and Fei-Fei, Li},
  booktitle={Proceedings of the IEEE conference on computer vision and pattern recognition},
  pages={4565--4574},
  year={2016}
}

@inproceedings{chen2020uniter,
  title={Uniter: Universal image-text representation learning},
  author={Chen, Yen-Chun and Li, Linjie and Yu, Licheng and Kholy, Ahmed El and Ahmed, Faisal and Gan, Zhe and Cheng, Yu and Liu, Jingjing},
  booktitle={European conference on computer vision},
  pages={104--120},
  year={2020},
  organization={Springer}
}

@inproceedings{li2022blip,
  title={BLIP: Bootstrapping language-image pre-training for unified vision-language understanding and generation},
  author={Li, Junnan and Li, Dongxu and Xiong, Caiming and Hoi, Steven},
  booktitle={International Conference on Machine Learning},
  pages={12888--12900},
  year={2022},
  organization={PMLR}
}

@article{lee2023aligning,
  title={Aligning text-to-image models using human feedback},
  author={Lee, Kimin and Liu, Hao and Ryu, Moonkyung and Watkins, Olivia and Du, Yuqing and Boutilier, Craig and Abbeel, Pieter and Ghavamzadeh, Mohammad and Gu, Shixiang Shane},
  journal={arXiv preprint arXiv:2302.12192},
  year={2023}
}

@misc{pinterest2024manas,
  title={Manas: Pinterest's ANN Infrastructure},
  author={{Pinterest Engineering}},
  year={2021},
  howpublished={\url{https://medium.com/pinterest-engineering/manas-hnsw-realtime-powering-realtime-embedding-based-retrieval-dc71dfd6afdd}}
}

@misc{pinterest2021searchsage,
  title={SearchSage: Learning Search Query Representations at Pinterest},
  author={{Pinterest Engineering}},
  year={2021},
  howpublished={\url{https://medium.com/pinterest-engineering/searchsage-learning-search-query-representations-at-pinterest-654f2bb887fc}}
}

@misc{pinlanding2025,
      title={PinLanding: Content-First Keyword Landing Page Generation via Multi-Modal AI for Web-Scale Discovery}, 
      author={Faye Zhang and Jasmine Wan and Qianyu Cheng and Jinfeng Rao},
      year={2025},
      eprint={2503.00619},
      archivePrefix={arXiv},
      primaryClass={cs.IR},
      url={https://arxiv.org/abs/2503.00619}, 
}



\end{document}